\def\year{2020}\relax
\documentclass[letterpaper]{article} 
\usepackage{aaai20}  
\usepackage{times}  
\usepackage{helvet} 
\usepackage{courier}  
\usepackage[hyphens]{url}  
\usepackage{graphicx} 
\usepackage{graphicx}  
\usepackage{amsmath}
\usepackage{multirow}

\title{End-to-end Learning of Object Motion Estimation from Retinal Events \\ for Event-based Object Tracking}
\author{Haosheng Chen\textsuperscript{\rm 1}, David Suter\textsuperscript{\rm 2}, Qiangqiang Wu\textsuperscript{\rm 1}, Hanzi Wang\textsuperscript{\rm 1}\thanks{The corresponding author.}\\ 
\textsuperscript{\rm 1}Fujian Key Laboratory of Sensing and Computing for Smart City,\\ School of Informatics, Xiamen University, China\\ 
\textsuperscript{\rm 2}School of Science, Edith Cowan University, Australia\\
\{haoshengchen, qiangwu\}@stu.xmu.edu.cn, d.suter@ecu.edu.au, hanzi.wang@xmu.edu.cn 
}
 
\begin{document}

\maketitle

\begin{abstract}
Event cameras, which are asynchronous bio-inspired vision sensors, have shown great potential in computer vision and artificial intelligence. However, the application of event cameras to object-level motion estimation or tracking is still in its infancy. The main idea behind this work is to propose a novel deep neural network to learn and regress a parametric object-level motion/transform model for event-based object tracking. To achieve this goal, we propose a synchronous Time-Surface with Linear Time Decay (TSLTD) representation, which effectively encodes the spatio-temporal information of asynchronous retinal events into TSLTD frames with clear motion patterns. We feed the sequence of TSLTD frames to a novel Retinal Motion Regression Network (RMRNet) to perform an end-to-end 5-DoF object motion regression. Our method is compared with state-of-the-art object tracking methods, that are based on conventional cameras or event cameras. The experimental results show the superiority of our method in handling various challenging environments such as fast motion and low illumination conditions. 
\end{abstract}

\section{Introduction}
\label{sec:Introduction}
\noindent Biological eyes are one of the most efficient and sophisticated neural systems. As a vital part of eyes, the retina can precisely and efficiently capture motion information \cite{murphy2018old}, especially for motions caused by moving objects \cite{olveczky2003segregation}, in natural scenes. Compared with the retina, most state-of-the-art motion estimation approaches are still limited by challenging conditions such as motion blur and high dynamic range (HDR) illuminations. There have been several attempts (e.g., \cite{mcintosh2016deep}) trying to imitate the retina by using artificial neural networks (ANNs). However, it is difficult for ANNs to imitate the asynchronous nature of the retina. In contrast, event cameras (e.g., DAVIS \cite{brandli2014240}) are asynchronous visual sensors with very high dynamic range and temporal resolution ($>$120 dB, $<$1 ms). These bio-inspired sensors help event-based methods to perform better in many computer vision and artificial intelligence tasks. In particular, event cameras can filter out non-motion information from the visual input, under stable illumination conditions or infrequent light variations: thus saving a lot of computation power, and giving clear clues about where object movement occurred.

\begin{figure}[t]
	\centering
	\resizebox{0.84\columnwidth}{!}{
	\includegraphics{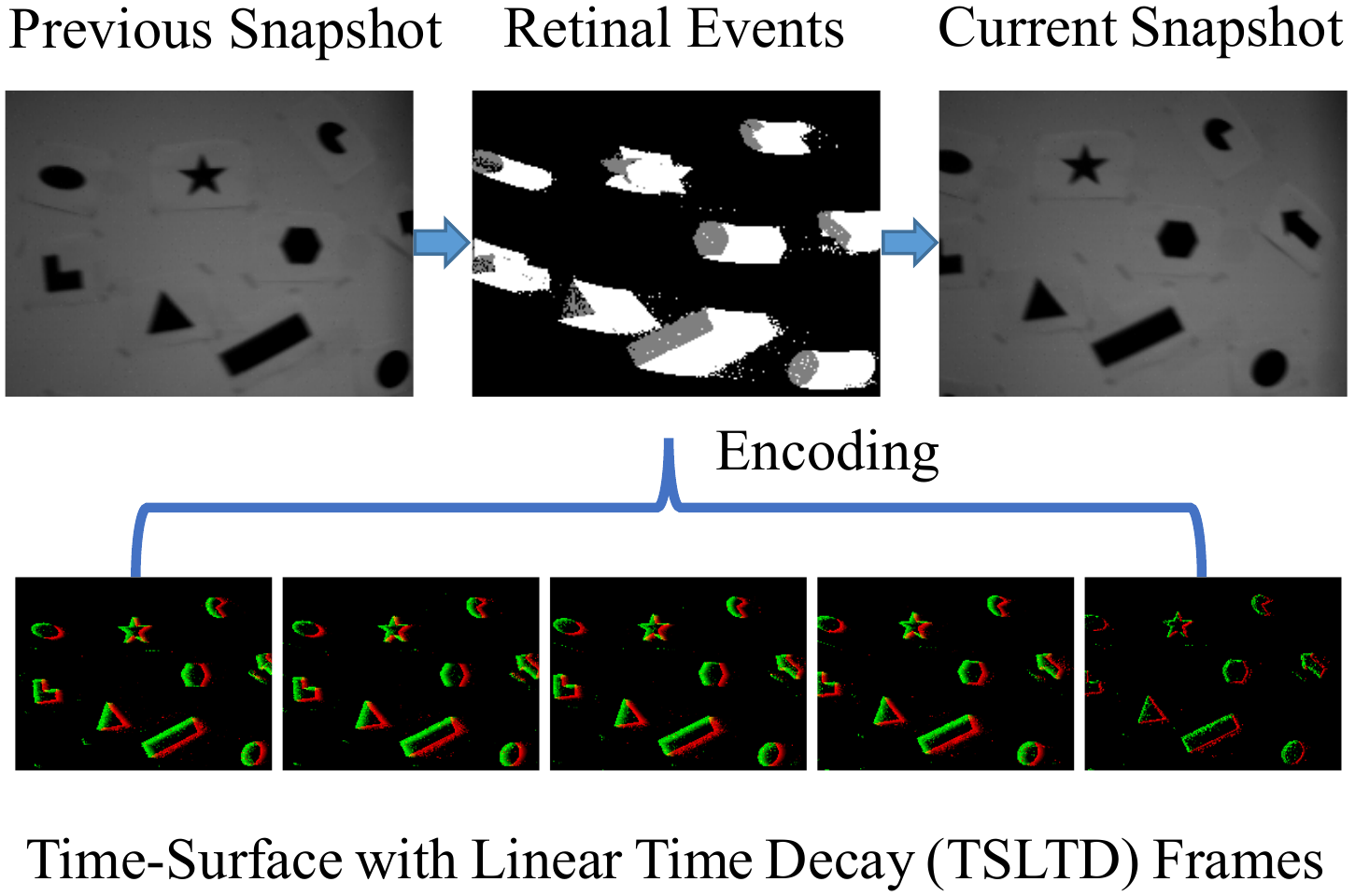}
    }
	\caption{An illustration of the encoding process of the proposed TSLTD frames. The left and right sub-figures in the first row show two snapshots, while the middle one shows the retinal events between the two snapshots. Note that the two snapshots are just for reference to show what a conventional camera would have seen before and after the retinal events occur. The second row shows a sequence of the encoded TSLTD frames of the retinal events.}
	\label{fig:retinalEvents}
\end{figure} 

Several 3D 6-DoF ego-motion estimation methods for event-based input, such as \cite{mueggler2014event,kim2016real,gallego2017event}, have been proposed during recent years. They have shown the superiority of event cameras on the motion estimation task. However, there are only a few studies devoted to analyzing object-level motion, and most of these studies are designed for some special scenarios (e.g., \cite{pikatkowska2012spatiotemporal} is for the pedestrian tracking scenario). Moreover, none of these methods are based on the regression methodology, which gives an explicit motion model for retinal events. As shown in Fig. \ref{fig:retinalEvents}, the retinal events, collected between the previous snapshot and the current snapshot, show clear visual patterns about object motions. With this intuition, we present a 5-DoF object-level motion estimation method based on the event camera inputs. 

In this study, we propose a novel deep neural network (called Retinal Motion Regression Network, abbreviated as RMRNet) to regress the corresponding 5-DoF object-level motion for moving objects. Here, the 5-DoF object-level motion is a 2D transform between the object bounding box in the previous frame and the estimated object bounding box in the current frame. The proposed RMRNet has a lightweight network structure, and it is end-to-end trainable. In order to leverage the proposed network, we encode asynchronous retinal events into a sequence of synchronous Time-Surface with Linear Time Decay (TSLTD) frames, as shown in the second row of Fig. \ref{fig:retinalEvents}. The TSLTD representation, based on the Time-Surface representation \cite{lagorce2017hots}, contains clear spatio-temporal motion patterns corresponding to the original retinal events, which is convenient for extracting motion information using RMRNet. Overall, this study makes the following contributions:

\begin{itemize}
\item We present the TSLTD representation that is amenable for preserving the spatio-temporal information of retinal events, and for training motion regression networks.
\item We introduce a 5-DoF object-level motion model to explicitly regress object motions for visual tracking.
\item We propose a Retinal Motion Regression Network (RMRNet) that allows one to end-to-end estimate the 5-DoF object-level motion, directly from TSLTD frames.
\end{itemize}

We evaluate our method on an event camera dataset and an extreme event dataset. The results demonstrate the superiority of our method when it is compared with several state-of-the-art object tracking methods.

\begin{figure*}[t]
	\centering
	\resizebox{1.75\columnwidth}{!}{
	\includegraphics{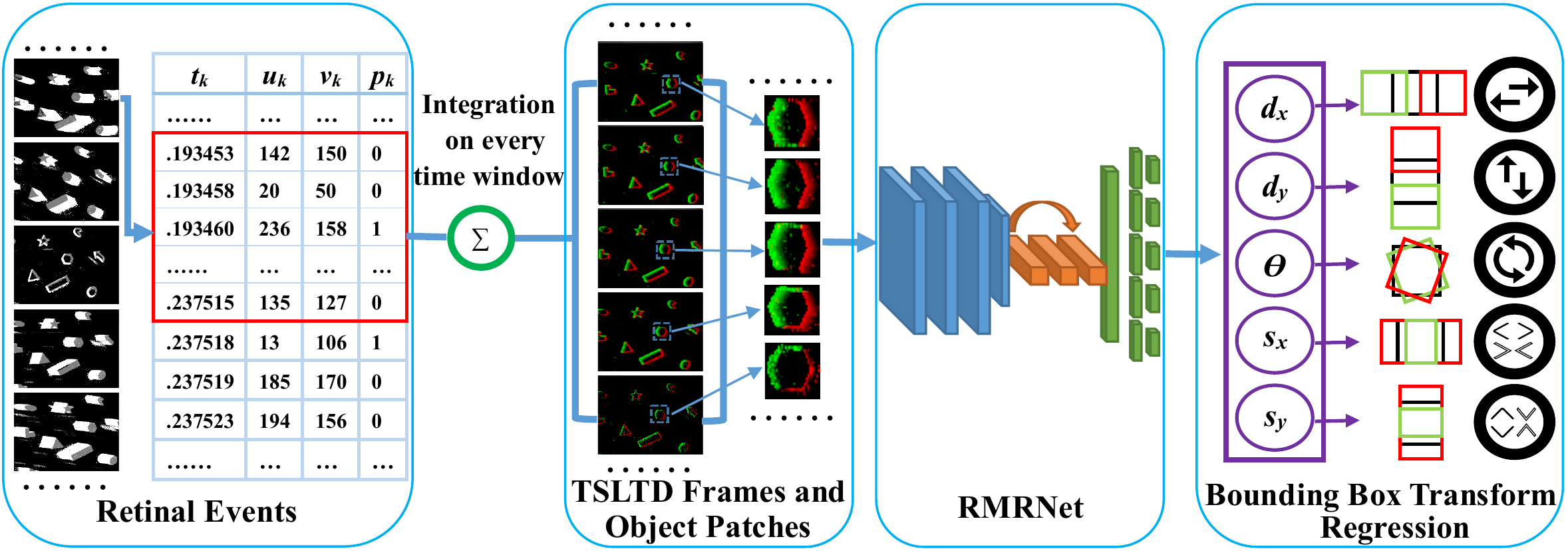}
    }
	\caption{The full pipeline of the proposed method. Initially, retinal events are integrated and formatted as a sequence of TSLTD frames. The input of RMRNet is the object patches cropped from the TSLTD frames for each object. These object patches are sent to a set of convolutional layers (marked in blue) to extract deep features. Then, these features are stacked or they are in one-by-one style, to pass through an LSTM module (marked in orange), for feature compression. Finally, the network is divided into five branches of fully connected layers (marked in green) to predict 5-DoF object-level motions separately.}
	\label{fig:pipeline}
\end{figure*}

\section{Related Work}
\label{sec:RelatedWork}
Event cameras have achieved great success on a variety of computer vision and artificial intelligence tasks \cite{gallego2019event}. Event-based motion estimation is a popular topic in these tasks. For 2D motion estimation, event-based optical flow can be calculated by using a sliding window variational optimization algorithm \cite{bardow2016simultaneous}, or the best point trajectories of the event data \cite{gallego2018unifying}, by using a self-supervised neural network \cite{Zhu_2019_CVPR}, or a time-slice block-matching method\cite{liu2018adaptive}. For 3D motion estimation, the event-based 6-DoF motion can be predicted by using a line constraint \cite{mueggler2014event}, using interleaved probabilistic filters \cite{kim2016real}, or using photometric depth maps \cite{gallego2017event}. For event-based visual-inertial data, the 3D motion information can also be estimated by using an extended Kalman filter on image features \cite{zihao2017event}, using a keyframe-based nonlinear optimization algorithm \cite{rebecq2017real} or using a continuous-time representation \cite{mueggler2018continuous}. As described in these works, event cameras have shown a unique and superior ability for the motion estimation task. From their results, we can see that event cameras usually outperform conventional cameras, especially when coping with some harsh conditions such as fast motion and HDR scenes.

Despite the fact that event-based object tracking methods can benefit a lot from the high spatio-temporal resolution and high HDR features of event cameras (when compared with conventional tracking methods e.g., \cite{fan2017robust,Yun_2017_CVPR,lan2018robust,Li_Zhu_Hoi_Song_Wang_Liu_2019,Qi_Zhang_Zhang_Su_Huang_Yang_2019,Huang_Zhou_2019}), there are only a few works done in this area. These works can be roughly divided into two categories. The works in the first category need a clustering process to group events into clusters. The works in the second category do not need the clustering process. 

In the first category, \cite{pikatkowska2012spatiotemporal} proposes a method for tracking multiple pedestrians with occlusions. They use a Gaussian mixture model for clustering. Similarly, \cite{camunas2017event} also rely on a clustering algorithm to track objects with occlusions using a stereo system of two event cameras. In \cite{glover2017robust}, they propose a variant of particle filter to track the cluster centers grouped by a Hough transform algorithm. 

In the second category, \cite{mitrokhin2018event} proposes an object detection and tracking method, which is built on top of the motion compensation concept, and they use the Kalman filter tracker for tracking. \cite{ramesh2018long} also proposes a detection and tracking method for long-term tracking using a local sliding window approach. \cite{barranco2018real} presents a real-time method for multi-target tracking based on both mean-shift clustering and Kalman tracking. 

For the above-mentioned studies, we have found that all of them involve handcrafted strategies. However, in our study, we prefer to learn an explicit 2D object motion model from original retinal events, with minimal human intervention, in an end-to-end manner.

\section{Proposed Method}
\label{sec:Methods}
Our method can directly predict frame-wise 5-DoF in-plane object-level motion (i.e., a 2D transform for bounding boxes) from retinal events. The full pipeline of our method is shown in Fig. \ref{fig:pipeline}. The retinal events are initially created by an event camera (we use event channel data from a DAVIS sensor \cite{brandli2014240}, as the input in this work), and the events are converted into a series of synchronous Time-Surface with Linear Time Decay (TSLTD) frames. Then the TSLTD frames are fed to our new Retinal Motion Regression Network (RMRNet) to regress the corresponding 5-DoF in-plane motions. In the remainder of this section, we introduce the proposed method in detail.

\subsection{Time-Surface with Linear Time Decay}
\label{subsec:RetinalEventFlow}
The $k$-th event $e_{k}$ of retinal events $\mathcal{E}$ can be represented as a quadruple:
\begin{equation}
e_{k} \doteq (u_{k}, v_{k}, p_{k}, t_{k}),
\end{equation}
where $u_{k}$ and $v_{k}$ are the horizontal and vertical coordinates of $e_{k}$; $p_{k}$ indicates the polarity (\emph{On} or \emph{Off}) of $e_{k}$, and $t_{k}$ is the timestamp of $e_{k}$. Retinal events can occur independently in an asynchronous manner, which makes it difficult for conventional computer vision algorithms to directly process the raw data. There are several attempts \cite{lagorce2017hots,sironi2018hats,maqueda2018event} that try to convert asynchronous retinal events to synchronous frames. For example, \cite{lagorce2017hots} adopts a hierarchical model, which is beneficial to the recognition task but containing less motion information. \cite{sironi2018hats,maqueda2018event} use specially designed histograms to format retinal events, which cut off the continuity of motion patterns in the temporal domain. In contrast, we prefer to create a clear and lightweight motion pattern for training our network with the help of event cameras, which allows our method to use a much smaller network structure, and maintain high precision estimation and real-time performance simultaneously.

In this work, we propose a synchronous representation of retinal events named Time-Surface with Linear Time Decay (TSLTD), as shown in Fig. \ref{fig:retinalEvents}. The TSLTD representation is based on the Time-Surface representation in \cite{lagorce2017hots}, but with two major differences: (1) the Time-Surface representation, which is designed for object recognition, consists of clear object contours with little motion information. We replace the exponential time decay kernel in Time-Surface with a linear time decay kernel to efficiently create effective motion patterns; (2) our TSLTD representation does not need the two hyper-parameters (i.e., the size of neighborhood and the time constant of exponential kernel) in Time-Surface. Thus, TSLTD can be effectively generalized to various target objects for motion estimation. 

In TSLTD, motion information, captured by an event camera, is encoded and represented as a set of TSLTD frames over every time window \emph{T}. The time window \emph{T} will be discussed later. Each of the TSLTD frames is initialized to a three-dimensional zero matrix $\mathcal{M} \in {\mathcal{N}^{h \times w \times 2}}$. Here $h$ and $w$ are the height and width of the event camera resolution, the third dimension indicates the polarity of events (and the information from \emph{On} or \emph{Off} events will be stored separately). Then asynchronous retinal events within the current time window are used to update the matrix in an ascending order in terms of their timestamps.

More specifically, supposing that we are processing a retinal event set $\mathcal{E}_{t_{s},t_{e}}$ which is collected between the start timestamp $t_s$ and the end timestamp $t_{e}=t_{s}+T$ of the current time window, to yield a new TSLTD frame $\mathcal{F}_{t_{s},t_{e}}$. $\mathcal{F}_{t_{s},t_{e}}$ is initialized to a 3D zero matrix $\mathcal{M}_{t_{s},t_{e}}$. During the updating process, we process ${{\cal E}_{t_{s},t_{e}}} = \left\{ {{e_i},{e_{i + 1}}, \ldots ,{e_j}} \right\}$ from ${e_i}$ to ${e_j}$, where ${e_i}$ is the first event and ${e_j}$ is the last event. Each of the events in $\mathcal{E}_{t_{s},t_{e}}$ triggers an update, which assigns a value ${g}$ to $\mathcal{M}_{t_{s},t_{e}}$ at the coordinates ${(u, v)}$ corresponding to the triggering event. The value ${g_k}$ for the $k$-th update caused by ${e_k}$ is calculated using the following equation:
\begin{equation}
{g_k} = round(255*\left( {{t_k} - {t_s}} \right)/T),
\end{equation}
where ${t_{k}}$ is the timestamp of ${e_{k}}$. So the assigned value for each update is proportional to the timestamp of the triggered event. ${t_k}-{t_s}$ is the linear time decay, and we use $255/T$ to normalize the decay. When a pixel of an object moves to a new coordinate, an event will occur at that coordinate and the TSLTD frame records a higher value of $g$ than the previous one at that coordinate. As a result, the time-motion variation information is naturally embedded in the TSLTD frames to form intensity gradients as shown in Fig. \ref{fig:pipeline}, which have shown a clear pattern of the direction and magnitude of the corresponding object motion. Therefore, the TSLTD format facilitates our network to extract the motion information through intensity gradient patterns that are embedded in the TSLTD frames.

There are two main problems that are related to the time window \emph{T} value, used in generating TSLTD frames. The first problem is that if \emph{T} is set to a large interval (e.g., more than 16 $ms$), there are two consequences for fast moving objects. Since these objects move fast, they can return to the previous position or move far from the previous position, during a large time interval. As a result, new motion patterns overlap and contaminate the previous patterns, or become too large to recognize. On the contrary, if \emph{T} is set to a small interval (e.g., less than 3 $ms$), TSLTD can only capture a very small movement, which may not be distinguished from sensor noises, and thus it may cause an ambiguous motion pattern (especially for a low-resolution event camera).  After testing \emph{T} with different objects and motions, we experimentally set \emph{T} to be $6.6$ $ms$ (according to the sampling frequency of 150 Hz) for good generalization performance.

\subsection{Network Structure}
\label{subsec:NetworkStructure}
Nowadays, pre-trained deep models, such as VGGNet \cite{simonyan2014very} and ResNet \cite{he2016deep}, are very popular among many computer vision tasks. But most of these pre-trained models were trained using three-channel RGB images, which is not the optimal option for two-channel TSLTD frames. In addition, since TSLTD frames have clear motion patterns, we do not need very deep and complex networks to extract very high-level features for general pattern recognition.

Here we design a lightweight network, named RMRNet, to learn object motions directly from TSLTD in an end-to-end manner. Between every two adjacent video frames, there are five TSLTD frames, which contain multiple objects. For individual object motion estimation, we crop object patches from the TSLTD frames. During the training stage, we crop the object patches from adjacent five TSLTD frames according to $\tau$ times of their axis-aligned bounding boxes, to preform a joint training. Here $\tau$ is a parameter that renders the cropped region slightly larger than the previous bounding box to capture the full pattern of object motion. During the test stage, we crop object patches frame-by-frame according to $\tau$ times of their axis-aligned bounding boxes. Finally, these object patches are resized to 64 $\times$ 64, and sent to the proposed RMRNet as the input. 

As shown in Fig. \ref{fig:pipeline}, the first part of RMRNet contains four convolutional layers for feature extraction. The initial three layers share the same kernel size of 3 $\times$ 3 with stride of 2, which is similar to the VGG-Network \cite{simonyan2014very}. The kernel size of the final layer, which is used to reduce the feature dimensions, is 1 $\times$ 1 with stride of 1. The filter numbers of the four layers are 32, 64, 128 and 32, respectively. The four convolutional layers are followed by a batch normalization layer. A dropout layer is added in the end during the training stage. Finally, the output feature is flattened and sent to the next part. The second part of RMRNet is an LSTM module. This module contains three layers with 1568 channels in each layer. By adding the LSTM module, we can stack object patches in one regression process. Then the LSTM module can fuse the stacked motion features from the CNN part to regress an accumulated motion, which is the motion between the first object patch and the final object patch. The final part of RMRNet is a set of fully connected layers, which are used to predict 5-DoF motions. The first fully connected layer has 1568 channels. Then the following layers are divided into five branches for different components of the 5-DoF motion. Each branch contains two layers, which respectively have 512 and 128 channels. This network structure is chosen due to its desirable performance on balancing both precision and speed. The output of RMRNet is a 5-DoF transform ($e_1$ to $e_5$), as described next.

\begin{table*}[t]
	\caption{The details of the ECD and EED datasets. The FM, BC, SO, HDR and OC, in the Challenges column, are fast motion,
		background clutter, small object, HDR scene and occlusion, respectively.}\smallskip
	\centering
	\resizebox{2\columnwidth}{!}{
		\begin{tabular}{|l|l|c|c|}
			\hline
			Dataset & Sequence names & Feature & Challenges\\
			\hline\hline
			ECD     & shapes\_translation   & B\&W shape objects mainly with translations    & FM \\
			ECD     & shapes\_6dof          & B\&W shape objects with various 6-DoF motions  & FM \\
			ECD     & poster\_6dof          & Natural textures with cluttered background and various 6-DoF motions  & FM+BC \\
			ECD     & slider\_depth         & Various artifacts at different depths with only translations  & BC  \\ \hline
			EED     & fast\_drone           & A fast moving drone under a very low illumination condition  & FM+SO+HDR   \\
			EED     & light\_variations     & Same with the upper one with extra periodical abrupt flash lights  & FM+SO+HDR   \\
			EED     & what\_is\_background  & A thrown ball with a dense net as foreground  & FM+OC   \\
			EED     & occlusions            & A thrown ball with a short occlusion under a dark environment  & FM+OC+HDR    \\
			\hline
		\end{tabular}
	}
    \label{tab:dataset}
\end{table*}

A 5-DoF transform ${\mathcal{T}_{i,j}^{o}}$ between a frame ${i}$ and the next frame ${j}$ for an object ${o}$ can be defined as a subset of the 2D affine transform on an object bounding box in the ${i}$-th frame:
\begin{equation}
{\mathcal{T}_{i,j}^{o}} \doteq (d_{x}, d_{y}, \theta, s_{x}, s_{y}),
\end{equation}
where ${\mathcal{T}_{i,j}^{o}}$ is represented as a quintet, $d_{x}$ and $d_{y}$ are respectively the horizontal and vertical displacement factors, $s_{x}$ and $s_{y}$ are respectively the horizontal and vertical scale factors, and $\theta$ is the rotation factor. Note that the rotation factor $\theta$ and the scaling factors $s_{x}$ and $s_{y}$ in ${\mathcal{T}_{i,j}^{o}}$ are ``in-place operations'', which means that we will keep center alignment before and after these two operations. The resulting coordinate transform is as follows:
\begin{equation}
\begin{bmatrix} {x}' \\ {y}' \end{bmatrix} =
\begin{bmatrix} s_{x} & 0 \\ 0 & s_{y} \end{bmatrix} \otimes
\begin{bmatrix} cos\theta & -sin\theta \\ sin\theta & cos\theta \end{bmatrix} \otimes
\begin{bmatrix} x + d_{x} \\ y + d_{y} \end{bmatrix}.
\end{equation}
Here the original coordinates $(x,y)$ of the bounding box of object ${o}$ in previous frame ${i}$ are transformed into the new coordinates $({x}',{y}')$ in current frame ${j}$ through the transform ${\mathcal{T}_{i,j}^{o}}$. The operator $\otimes$ indicates an in-place operation. Note that the five parameters $e_1$ to $e_5$ predicted by RMRNet are normalized to a range of $[-1.0,1.0]$ using the Tanh activation function. Then we set five boundary parameters $p_1$ to $p_5$ for $e_1$ to $e_5$ to constrain the range of object movements. Finally, the transform ${\mathcal{T}_{i,j}^{o}}$ is calculated as follows:
\begin{equation}
\label{eq:5dof}
{\cal T}_{i,j}^o = \left\{ {\begin{array}{*{20}{l}}
	{{d_x} = {e_1}*{p_1}}\\
	{{d_y} = {e_2}*{p_2}}\\
	{\theta  = ({e_3}*{p_3})*\pi /180}\\
	{{s_x} = 1 + {e_4}*{p_4}}\\
	{{s_y} = 1 + {e_5}*{p_5}}\\
	\end{array}} \right..
\end{equation}
In this paper, we respectively fix $p_1$ to $p_5$ to 72, 54, 30, 0.2 and 0.2, and fix $\tau$ to 1.2 according to the 240 $\times$ 180 resolution of the DAVIS camera for RMRNet. This parameter setting will allow a relatively large range for object movements. Thus, the setting is suitable for estimating the majority of object motions, which includes most of the fast movements.

\subsection{Learning Approach}
\label{subsec:LearningApproach}
There are two key points that should be mentioned in relation to the training stage. The first one is that if we only use B\&W object samples (i.e., black objects with a white background) as the training data, our network can only learn some relatively simple motion patterns for the corresponding object motions. Thus, we follow the standard five-fold cross-validation protocol and use the object pairs (refer to the next section) from the \emph{shapes\_6dof}, \emph{poster\_6dof} and \emph{light\_variations} sequences as the training and validation data (while the other five sequences are unseen during the training stage) to train and validate the proposed RMRNet. The second point is that it is difficult to learn an object motion model from a single TSLTD frame. There are five TSLTD frames for every object pair. A single TSLTD frame usually includes only a small movement and it has only a weak motion pattern. After extensive experiments, we find that stacking five TSLTD frames in one prediction has gained the optimal performance during the training stage.

The proposed network is trained using the ADAM solver with a mini-batch size of 16. We use a fixed learning rate of 0.0001, and our loss function is the MSE loss:
\begin{equation}
MSE_{loss}\left( {\hat {\cal T},{\cal T}} \right) = \frac{1}{{{N_{train}}}}\sum\limits_{l = 1}^{{N_{train}}} {{{\left\| {{{\hat {\cal T}}_l} - {{\cal T}_l}} \right\|}^2}},
\end{equation}
where $\hat{\cal{T}}$ is the estimated 5-DoF motion, $\cal{T}$ is the corresponding ground truth, $N_{train}$ is the number of training samples, and the subscript $l$ indicates the $l$-th sample.

\section{Experiments}
\label{sec:Experiments}

\subsection{Pre-processing}
\label{subsec:Pre-processing}
For our evaluation, we use a challenging mixed event dataset including a part of the popular Event Camera Dataset (ECD) \cite{mueggler2017event} and the Extreme Event Dataset (EED) \cite{mitrokhin2018event}, which were recorded using a DAVIS event camera in real-world scenes. The details of the dataset can be found in Table \ref{tab:dataset}. Note that the mixed dataset contains both the event data sequences and the corresponding video sequences for every sequence. Since the ECD dataset does not provide ground truth bounding boxes for object tracking, we labeled a rotated or an axis-aligned rectangle bounding box as the ground truth for each object in the dataset to evaluate all the competing methods.

We choose five state-of-the-art object tracking methods, including SiamFC \cite{bertinetto2016fully}, ECO \cite{danelljan2017eco}, SiamRPN++ \cite{Li_2019_CVPR}, ATOM \cite{Danelljan_2019_CVPR} and E-MS \cite{barranco2018real}, as our competitors. About these competitors: SiamFC and SiamRPN++ are fast and accurate methods, which are based on Siamese networks. ECO and ATOM are state-of-the-art methods that have achieved great performance on various datasets. E-MS \cite{barranco2018real} is recently proposed for event-based target tracking based on mean-shift clustering and Kalman filter. We extend E-MS to support bounding box-based tracking, by employing the minimum enclosing rectangle of those events that belong to an identical cluster center as its estimated bounding box. 

To perform ablation studies, we also compare our RMRNet with an event-based variant of ECO (called as ECO-E) and a variant of RMRNet (called as RMRNet-TS). ECO-E, which uses our proposed TSLTD frames as its inputs, is an event-based variant of ECO \cite{danelljan2017eco}. ECO-E is used to evaluate the performance of ECO on the event data sequences. RMRNet-TS uses the classical Time-Surface frames in Hots \cite{lagorce2017hots} instead of using the proposed TSLTD frames as its inputs. RMRNet-TS is used to evaluate the performance of the classical Time-Surface representation and our TSLTD representation. For SiamFC, ECO, SiamRPN++ and ATOM, we use the video sequences as their inputs. For ECO-E, E-MS, RMRNet-TS and the proposed RMRNet, we use the event data sequences (in quadruple format) as their inputs. For all the five competitors, we use their released codes, default parameters and best pre-trained models.

\begin{table*}[t]
	\caption{Results obtained by the competitors and our method on the ECD dataset. The best results are in \textbf{bold}.}\smallskip
	\label{tab:ecd_results}
	\centering
	\resizebox{2\columnwidth}{!}{
		\begin{tabular}{|c|c|c|c|c|c|c|c|c|}
			\hline
			\multirow{2}{*}{Method} & \multicolumn{2}{c|}{shapes\_translation} & \multicolumn{2}{c|}{shapes\_6dof}   & \multicolumn{2}{c|}{poster\_6dof}   & \multicolumn{2}{c|}{slider\_depth} \\ \cline{2-9}
			& AOR                 & AR                 & AOR              & AR               & AOR              & AR               & AOR                & AR            \\ \hline\hline
			SiamFC\cite{bertinetto2016fully}                  & 0.812             & 0.940            & 0.835          & 0.968          & 0.830          & 0.956          & 0.909            & \textbf{1.000}       \\ \hline
			ECO\cite{danelljan2017eco}                     & 0.823               & 0.943              & 0.847            & 0.969            & 0.846            & 0.960            & \textbf{0.947}              & \textbf{1.000}         \\ \hline
			SiamRPN++\cite{Li_2019_CVPR}     & 0.790             & 0.942            & 0.779          & 0.972          & 0.753          & 0.899          & 0.907            & \textbf{1.000}       \\ \hline
			ATOM\cite{Danelljan_2019_CVPR}       & 0.815             & 0.945            & 0.803          & 0.974          & 0.835          & 0.961          & 0.897            & \textbf{1.000}       \\ \hline
			ECO-E\cite{danelljan2017eco}                     & 0.821               & 0.941              & 0.834            & 0.960            & 0.783            & 0.878            & 0.771              & 0.993         \\ \hline
			E-MS\cite{barranco2018real}                  & 0.675    & 0.768   & 0.612 & 0.668 & 0.417 & 0.373 & 0.447   & 0.350  \\ \hline
			RMRNet-TS                  & 0.491    & 0.564   & 0.467 & 0.509 & 0.504 & 0.558 & 0.814   & 0.993  \\ \hline
			RMRNet                  & \textbf{0.836}    & \textbf{0.951}   & \textbf{0.866} & \textbf{0.980} & \textbf{0.859} & \textbf{0.962} & 0.915   & \textbf{1.000}  \\ \hline
		\end{tabular}
    }
\end{table*}

\begin{table*}[t]
	\caption{Results obtained by the competitors and our method on the EED dataset. The best results are in \textbf{bold}.}\smallskip
	\label{tab:eed_results}
	\centering
	\resizebox{2\columnwidth}{!}{
		\begin{tabular}{|c|c|c|c|c|c|c|c|c|}
			\hline
			\multirow{2}{*}{Method} & \multicolumn{2}{c|}{fast\_drone} & \multicolumn{2}{c|}{light\_variations} & \multicolumn{2}{c|}{what\_is\_background} & \multicolumn{2}{c|}{occlusions} \\ \cline{2-9}
			& AOR               & AR           & AOR                & AR                & AOR                   & AR                & AOR              & AR           \\ \hline\hline
			SiamFC\cite{bertinetto2016fully}                  & 0.766           & \textbf{1.000}      & 0.772            & \textbf{0.947}           & 0.712               & 0.833          & 0.148          & 0.000          \\ \hline
			ECO\cite{danelljan2017eco}                     & 0.830           & \textbf{1.000}      & 0.782            & 0.934           & 0.675               & 0.750           & 0.209          & 0.333          \\ \hline
			SiamRPN++\cite{Li_2019_CVPR}     & 0.717           & 0.941      & 0.497            & 0.500           & 0.653      & 0.833      & 0.096          & 0.167      \\ \hline
			ATOM\cite{Danelljan_2019_CVPR}     & 0.763           & \textbf{1.000}      & 0.652            & 0.921           & \textbf{0.725}               & \textbf{0.917}               & 0.387          & 0.500          \\ \hline
			ECO-E\cite{danelljan2017eco}                     & 0.728           & 0.882      & 0.685            & 0.803           & 0.099               & 0.000           & 0.308          & 0.333          \\ \hline
			E-MS\cite{barranco2018real}                  & 0.313  & 0.307 & 0.325   & 0.321  & 0.362               & 0.360           & 0.356 & 0.353 \\ \hline
			RMRNet-TS                  & 0.199  & 0.118 & 0.096   & 0.066  & 0.108               & 0.000           & 0.000 & 0.000 \\ \hline
			RMRNet                  & \textbf{0.892}  & \textbf{1.000} & \textbf{0.802}   & \textbf{0.947}  & 0.202               & 0.083           & \textbf{0.716} & \textbf{0.833} \\ \hline
		\end{tabular}
    }
\end{table*}

The output of our RMRNet is a 2D 5-DoF frame-wise bounding box transform model, estimated between two adjacent frames. In order to evaluate the quality of the estimated frame-wise bounding box transform model, we compare the proposed RMRNet with all the competitors on frame-wise object tracking, that is, our evaluation on these competing methods is based on object pairs, each of which includes two object regions on two adjacent frames corresponding to an identical object. During the evaluation, we treat each of the object pairs as a tracking instance in the corresponding frame. In addition, all the competitors only estimate axis-aligned bounding boxes, whereas the proposed RMRNet can estimate both rotated and axis-aligned bounding boxes (by using the 5-DoF motion model). Therefore, we evaluate all the methods on the axis-aligned ground truth bounding boxes for a fair precision comparison.

\subsection{Evaluation Metrics}
\label{subsec:EvaluationMetrics}
For evaluating the precision of all the methods, we calculate the Average Overlap Rate (AOR) as follow:
\begin{equation}
\label{eq:aor}
{AOR} = \frac{1}{{{N_{rep}}}}\frac{1}{{{N_{pair}}}}\sum\limits_{u = 1}^{{N_{rep}}} {\sum\limits_{v = 1}^{{N_{pair}}} {\frac{{O_{u,v}^E \cap O_{u,v}^G}}{{O_{u,v}^E \cup O_{u,v}^G}}}},
\end{equation}
where $O_{u,v}^{E}$ is the estimated bounding box in the $u$-th round of the cross-validation for the $v$-th object pair, and $O_{u,v}^{G}$ is the corresponding ground truth. Eq. (\ref{eq:aor}) shows that the AOR measure is related to the Intersection over Union (IoU). ${N_{rep}}$ is the repeat times of the cross-validation, and ${N_{pair}}$ is the number of object pairs in the current sequence. We set ${N_{rep}}$ to 5 for all the following experiments. 

We also calculate the Average Robustness (AR) to measure the robustness of all the competing methods as follow:
\begin{equation}
{AR} = \frac{1}{{{N_{rep}}}}\frac{1}{{{N_{pair}}}}\sum\limits_{u = 1}^{{N_{rep}}} {\sum\limits_{v = 1}^{{N_{pair}}} {succes{s_{u,v}}}},
\end{equation}
where $succes{s_{u,v}}$ indicates that whether the tracking in the $u$-th round for the $v$-th pair is successful or not (0 means failure and 1 means success). If the AOR value obtained by a method for one object pair is under 0.5, we will consider it as a tracking failure case.

\begin{figure*}[t]
	\centering
	\resizebox{1.53\columnwidth}{!}{
	\includegraphics{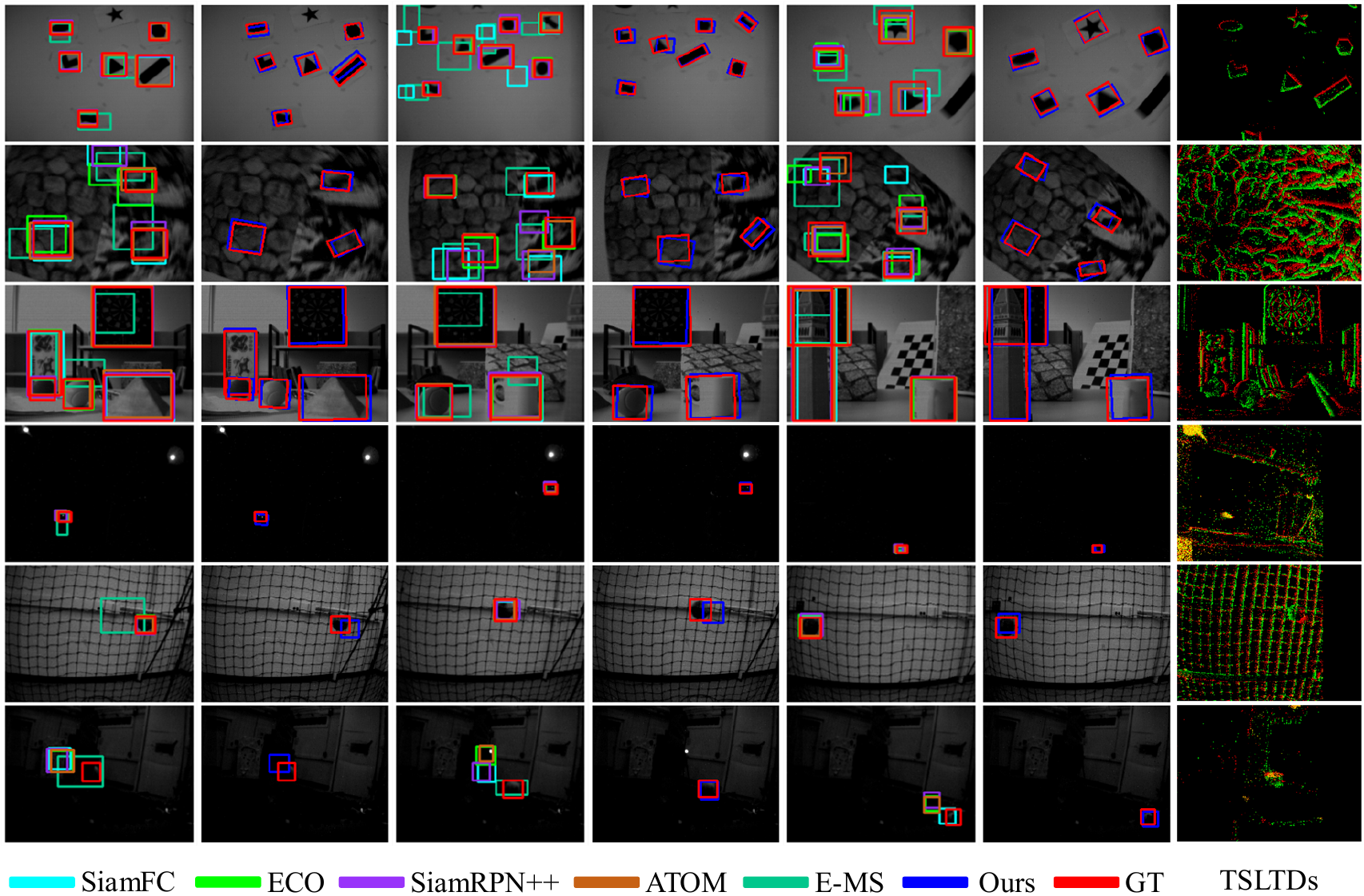}
    }
	\caption{Tracking results obtained by SiamFC, ECO, SiamRPN++, ATOM, E-MS and our method. Each row represents a sequence of the two datasets. From top to bottom, the corresponding sequences are \emph{shape\_6dof}, \emph{poster\_6dof}, \emph{slider\_depth}, \emph{light\_variations}, \emph{what\_is\_background} and \emph{occlusions}, respectively. From left to right, the first, third and fifth columns show the results of the competing methods with the axis-aligned GT. The second, fourth and sixth columns show the results of our method with the rotated GT. The seventh column show the actual TSLTD frames of the second column. Best viewed in color.}
	\label{fig:results}
\end{figure*}

\subsection{Evaluation on the Event Dataset}
\label{subsec:Evaluation}
We use the mixed event dataset to evaluate the eight competing methods. We choose the \emph{shapes\_translation}, \emph{shapes\_6dof}, \emph{poster\_6dof} and \emph{slider\_depth} sequences from the ECD dataset \cite{mueggler2017event} as the representative sequences for comparison. The first three sequences have increasing motion speeds. The fourth sequence has a constant motion speed. The object textures of these sequences vary from simple B\&W shapes to complicated artifacts. For these sequences, we are mainly concerned with the performance of all methods for various motions, especially for fast 6-DoF motion, and for different object shapes and textures.

For comparison, the quantitative results are reported in Table \ref{tab:ecd_results}. We also provide some representative qualitative results obtained by SiamFC, ECO, SiamRPN++, ATOM, E-MS and our method in the top three rows of Fig. \ref{fig:results}. From Table \ref{tab:ecd_results}, we can see that our method achieves the best performance on the first three sequences and it achieves the second best performance on the fourth sequence. SiamFC, ECO, SiamRPN++ and ATOM also achieve competitive results. However, as we can see in Fig. \ref{fig:results}, our method has achieved better performance in estimating fast motion. In comparison, SiamFC, ECO, SiamRPN++ and ATOM usually lose the tracked objects, due to the influence of motion blur. Comparing with the original ECO, ECO-E has achieved inferior performance, which shows that state-of-the-art object tracking methods, like ECO, are not suitable to be directly applied to event data. The classical Time-Surface frames are designed for object recognition and detection, which contain less motion information for motion estimation and object tracking. Thus, RMRNet-TS shows much inferior results in this evaluation. By leveraging the high temporal resolution feature of the event data, E-MS can also effectively handle most fast motions. However, E-MS is less effective to handle complicated object textures and cluttered backgrounds (e.g., for the \emph{poster\_6dof} and \emph{slider\_depth} sequences). In contrast, the proposed RMRNet outperforms E-MS by a large margin, which shows the superiority of our method in handling various object textures and cluttered backgrounds.

Moreover, we also choose the EED dataset to evaluate the eight competing methods. The EED dataset \cite{mitrokhin2018event} contains four challenging sequences: \emph{fast\_drone}, \emph{light\_variations}, \emph{occlusions} and \emph{what\_is\_background}. The first three sequences respectively record a fast moving drone under low illumination environments. The fourth sequence records a moving ball with a net as foreground. Using this dataset, we want to evaluate the competing methods in low illumination conditions and in occlusion situations.

The quantitative results and some representative qualitative results are shown in Table \ref{tab:eed_results} and Fig. \ref{fig:results}, respectively. From the results, we can see that our method achieves the best performance on most sequences except for the \emph{what\_is\_background} sequence, on which our method has obtained low AOR and AR. This is because that the foreground net in the \emph{what\_is\_background} sequence covers the ball, which destroys the corresponding motion patterns. In contrast, our method has achieved the highest AOR and AR on the \emph{occlusions} sequence. This is because that in this sequence, the occlusion time is short and only one object pair involves the occlusion. Our method fails at tracking that object pair but it successfully estimates the other object pairs. As the competitors, SiamFC, ECO, SiamRPN++ and ATOM show their low effectiveness with fast motion and low illumination conditions. Moreover, ECO-E and RMRNet-TS are respectively inferior to ECO and RMRNet. Although SiamFC and ECO achieve relatively good results for the first two sequences, the sequences include relatively clean backgrounds, which helps the two methods to achieve the performance. However, if we add a small amount of noise around the objects, the performance of the two methods will be significantly reduced. In contrast, our method can maintain its performance even with the severe sensor noises of an event camera, as shown in the seventh column of Fig. \ref{fig:results}. Meanwhile, the performance of E-MS is highly affected by the sensor noises in the HDR environments. Thus, E-MS show unsatisfied results on the EED dataset.

\subsection{Time Cost}
\label{subsec:timecost}
Since the proposed RMRNet is a relatively shallow network, our method has an advantage of relatively high efficiency. The proposed RMRNet is implemented using PyTorch on a PC with an Intel i7 CPU and an NVIDIA GTX 1080 GPU. For the mixed dataset, our method achieves real-time performance and the average computational time for each object pair (including five TSLTD frames) is 38.57 ms. 

\section{Conclusion}
\label{sec:Conclusion}
In this paper, we demonstrate the great potential of event cameras for object tracking under severe conditions such as fast motion and low illumination scenes. By using the event camera, our method only extracts motion related information from the inputs. Then we present the TSLTD representation to encode input retinal events into a sequence of synchronous TSLTD frames. TSLTD can represent the spatio-temporal information with clear motion patterns. Finally, to leverage the motion clues contained in TSLTD, we propose the RMRNet to regress 5-DoF motion information in an end-to-end manner. Extensive experiments demonstrate the superiority of our method over several other state-of-the-art object tracking methods.

\section{Acknowledgments}
This work is supported by the National Natural Science Foundation of China under Grants U1605252 and 61872307. Haosheng Chen and David Suter acknowledge funding from ECU that enabled the first author to visit ECU and undertake a small portion of this work.

\bibliographystyle{aaai}
\bibliography{references}

\end{document}